\documentclass[pdflatex,sn-mathphys-num]{sn-jnl}

\usepackage{graphicx}%
\usepackage{multirow}%
\usepackage{amsmath,amssymb,amsfonts}%
\usepackage{amsthm}%
\usepackage{mathrsfs}%
\usepackage[title]{appendix}%
\usepackage{xcolor}%
\usepackage{textcomp}%
\usepackage{manyfoot}%
\usepackage{booktabs}%
\usepackage{algorithm}%
\usepackage{algorithmicx}%
\usepackage{algpseudocode}%
\usepackage{listings}%

\theoremstyle{thmstyleone}%
\theoremstyle{thmstyletwo}%

\theoremstyle{thmstylethree}%

\begin{document}

\title[Article Title]{Learning from the Unseen: Generative Data Augmentation for Geometric-Semantic Accident Anticipation}


\author[1,2]{\fnm{Yanchen} \sur{Guan}}

\author[1,3]{\fnm{Haicheng} \sur{Liao}}

\author[1,2]{\fnm{Chengyue} \sur{Wang}}
\author[1,3]{\fnm{Xingcheng} \sur{Liu}}
\author[1,2]{\fnm{Jiaxun} \sur{Zhang}}
\author[4]{\fnm{Keqiang} \sur{Li}}
\author*[1,2,3]{\fnm{Zhenning} \sur{Li}}

\affil[1]{\orgdiv{State Key Laboratory of Internet of Things for Smart City}, \orgname{University of Macau}, \orgaddress{ \city{Macau SAR}, \postcode{999078},  \country{China}}}

\affil[2]{\orgdiv{Department of Civil Engineering}, \orgname{University of Macau}, \orgaddress{ \city{Macau SAR}, \postcode{999078},  \country{China}}}

\affil[3]{\orgdiv{Department of Computer and Information Science}, \orgname{University of Macau}, \orgaddress{ \city{Macau SAR}, \postcode{999078},  \country{China}}}

\affil[4]{\orgdiv{Department of Automotive Engineering}, \orgname{Tsinghua University}, \orgaddress{ \city{Beijing}, \postcode{100084},  \country{China}}}



\abstract{
Anticipating traffic accidents is a critical yet unresolved problem for autonomous driving, hindered by the inherent complexity of modeling interactions between road users and the limited availability of diverse, large-scale datasets. To address these issues, we propose a dual-path framework. On the one hand, we employ a video synthesis pipeline that, guided by structured prompts, derives feature distributions from existing corpora and produces high-fidelity synthetic driving scenes consistent with the statistical patterns of real data. On the other hand, we design a graph neural network enriched with semantic cues, enabling dynamic reasoning over both spatial and semantic relations among participants.
To validate the effectiveness of our approach, we release a new benchmark dataset containing standardized, finely annotated video sequences that cover a broad spectrum of regions, weather, and traffic conditions. Evaluations across existing datasets and our new benchmark confirm notable gains in both accuracy and anticipation lead time, highlighting the capacity of the proposed framework to mitigate current data bottlenecks and enhance the reliability of autonomous driving systems.
}




\maketitle

\section{Introduction}
In recent years, autonomous driving technologies have progressed rapidly, and self-driving vehicles are increasingly regarded as promising solutions for enhancing road safety \cite{bagloee2016autonomous,kalra2016driving,li2024steering}. Anticipating potential accidents is a key component in ensuring timely risk awareness within such scenarios, as it focuses on forecasting potential collisions from dashcam recordings and issuing early alerts. This ability allows drivers or automated systems to take preventive actions in advance, thereby lowering accident likelihood and improving traffic safety \cite{fang2023vision, jiang2021event, liu2020driver}.  

Distinct from generic time-series prediction tasks, accident anticipation is characterized by its reliance on anomalous interactions among traffic participants within specific environments, rather than merely long-term sequential dynamics. Existing studies mainly emphasize extending temporal modeling, thereby improving general sequence understanding but overlooking the underlying causes of accidents \cite{wang2024dc,guan2025domain}. This often results in weakly perceived interactions among distant targets, while close interactions suffer from extremely short time-to-accident (ToA), occlusions, or blurred cues—issues that are particularly detrimental in rapidly evolving traffic scenes \cite{cheong2017reflection}.

Furthermore, the lack of sufficiently large and reliable datasets is also a primary obstacle. Since traffic crashes occur infrequently and in unpredictable circumstances, their randomness and time-critical nature make the collection, curation, and labeling of accident videos particularly costly and labor-intensive \cite{sadeky2010real}. Consequently, accident datasets are limited in size and vary substantially across sources, leading to instability in model performance and a heightened risk of overfitting when evaluated on small, independent datasets \cite{wang2023gsc}. Moreover, cross-dataset discrepancies are pronounced, for example, the DAD dataset \cite{chan2017anticipating} reflects subtropical urban scenes without snowfall, while the CCD dataset \cite{BaoMM2020} primarily captures traffic scenarios on temperate open roads. Such differences may cause systematic biases, where specific scenarios or categories are underrepresented.

To tackle these challenges, we propose a dual-pronged pipeline. We first integrate a frozen video generation module into the processing pipeline of the original data, leveraging the environmental feature distribution of the raw dataset to construct new traffic scenarios. This approach increases both the volume and diversity of training data while preserving the intrinsic characteristics of the dataset to the greatest extent. Subsequently, to overcome limitations of prior research, we propose a semantic and geometric enhanced dynamic graph convolutional network for accident anticipation inference. This architecture captures complex behavioral interactions among participants through joint semantic and spatial-visual fusion, adaptively weighting critical targets to guide the model’s focus toward hazardous interactions. Furthermore, instead of treating frames independently, our model processes short video segments, effectively enlarging the receptive field, alleviating transient information loss, and enhancing prediction accuracy. 

For empirical validation, we present the Multi-source Accident Anticipation (MAA) dataset, a new and challenging real-world benchmark with richer accident samples and more detailed annotations than prior datasets. We further replicate several representative models based on RNNs, GCNs, Transformers, and self-supervised approaches on the MAA dataset. Results indicate that our framework consistently surpasses prior art across diverse datasets. Moreover, the proposed augmentation strategy effectively mitigates model performance bottlenecks, leading to more stable accident anticipation.

Our main contributions can be summarized as follows: 
\begin{itemize} 

\item We present a traffic video synthesis pipeline for data augmentation in accident anticipation. The method produces diverse driving scenarios while maintaining the statistical characteristics of the original environments, thereby alleviating the challenge of limited training data.

\item We develop a semantic and geometric enhanced dynamic graph convolutional network with temporal convolutional layers. By integrating the behavioral semantics and visual information of traffic participants with an enhanced temporal reasoning module, our framework achieves more accurate accident anticipation in multiple experiments.

\item We provide the MAA dataset, which constitutes the most extensive collection of accident cases with detailed annotations, supporting reproducible evaluation and future exploration.  \end{itemize}

\section{Related Work}
\label{sec:Related Work}
Accident anticipation aims to forecast collisions before they occur and to provide early warnings, allowing proactive safety measures to be taken. The concept was first explored by Chan et al. \cite{chan2017anticipating}, who introduced a Dynamic Spatial Attention framework using egocentric visual input, demonstrating the potential of predicting accidents directly from dashcam footage. Subsequent advances extended this idea: Karim et al. \cite{karim2022dynamic} incorporated temporal dependencies through a Dynamic Spatio-Temporal Attention model, while others introduced graph-based approaches that represent vehicles and pedestrians as nodes, with edges encoding their interactions \cite{fatima2021global,song2024dynamic,chen2021graph,ou2020data}. Such methods often rely on dense connectivity to approximate relational structures \cite{wang2023gsc,fang2023vision, thakur2024graph}. Parallel efforts have explored incorporating motion cues such as optical flow and depth information to refine object perception, which further improves accident anticipation accuracy \cite{liu2023net,liao2024real}. Despite these developments, key challenges persist, including the inefficiency of relation modeling and the limited scale of available datasets. Graph structures constructed solely from visual information struggle to capture real-world traffic semantics such as road constraints and vehicle priority relationships, and they often exhibit instability in rare scenarios \cite{liang2020learning,mohamed2020social}. Moreover, the limited scale of existing public datasets further exacerbates the long-tailed distribution of rare events, resulting in weak model generalization capability \cite{gesnouin2022assessing,martin2019drive}.

Beyond structured interaction modeling, research on scene graph generation has shown that explicitly representing semantic relations between objects can enrich video interpretation \cite{wu2020comprehensive,xiao2023review}. In addition, recognizing human actions from video provides a pathway to deeper understanding of ongoing events. Vision–language models (VLMs) have become a promising tool to bridge visual signals with textual representations \cite{zhang2024vision,radford2021learning}. These pretrained systems bring zero-shot interpretability, enhancing both contextual reasoning and content recognition \cite{lohner2024enhancing}. In video analysis specifically, VLMs exploit temporal continuity across frames, enabling them to capture evolving actions and mitigate the limitations of single-frame inference, thereby converting continuous visual information into semantic information \cite{gao2024clip}. Compared with fragmented frame-level visual cues, such continuous semantic information provides a more coherent description of inter-agent interactions and critical environmental influences \cite{fang2021clip2video,zellers2021merlot}. In addition, the zero-shot reasoning capability of VLMs enables the extraction of diverse semantic attributes for each sample without relying on multiple specialized models \cite{zhang2024vision}. This facilitates a unified characterization of dataset-wide semantic distributions and offers valuable guidance for constructing semantically aligned synthetic data \cite{bar2022text2live,wendland2022generation}.

\begin{figure}[t]
\centering  
\includegraphics[width=1.0\textwidth]{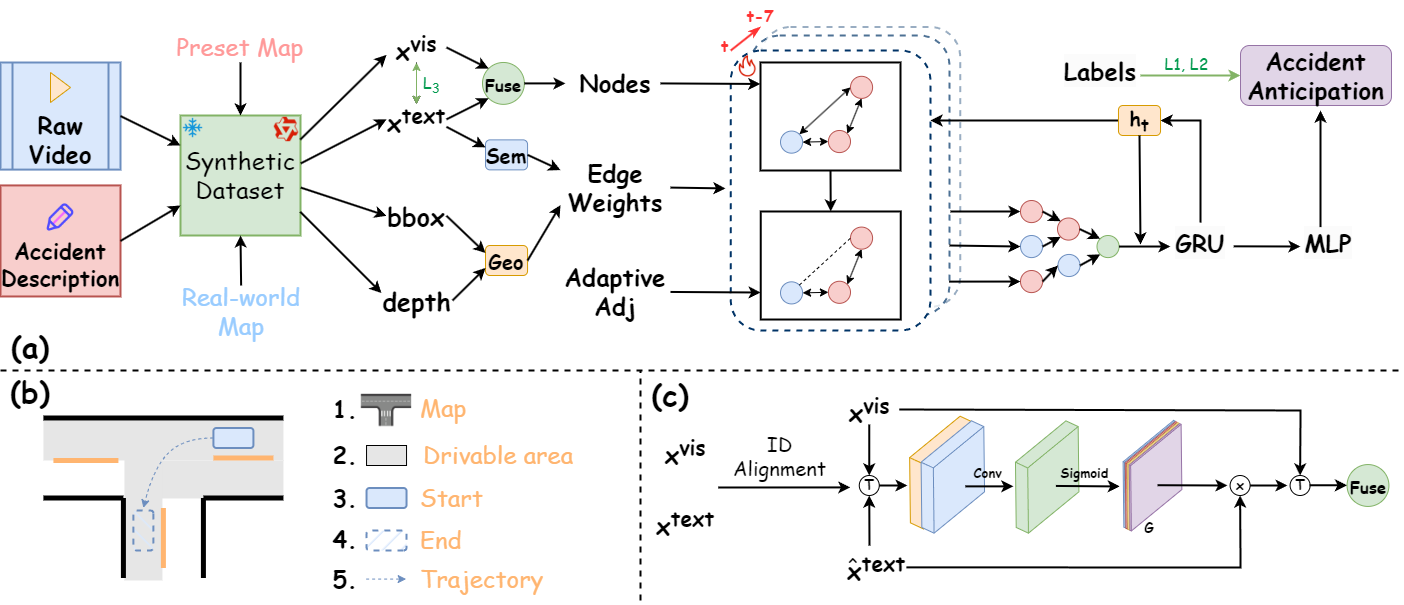}
\caption{The overall framework of our proposed model. (a) The process from raw video to accident anticipation results including traffic video generation and accident anticipation. (b) The important factors that constitute the traffic scene and the order in which they are obtained. (c) Semantic feature and visual feature fusion module.
}
\label{Overall}
\end{figure}

\section{Method}
\label{sec:Methodology}

\subsection{Problem Setup}
This work aims to determine, from a stream of dashcam frames, how the risk of a
traffic accident evolves over time and to identify the moment at which such risk
becomes critical.  Let $\mathcal{I}=\{i_{1}, i_{2},\ldots,i_{L}\}$ denote the
input video sequence consisting of $L$ frames.  At each frame index $m$, the
proposed model produces a risk estimate, yielding a probabilistic profile
$\mathcal{U}=\{u_{1},u_{2},\ldots,u_{L}\}$, where $u_{m}$ indicates the inferred
accident likelihood at that instant.

An accident-warning decision is triggered once the predicted score surpasses a
decision level $\delta$ for the first time at index $\hat{m}$, that is,
$u_{\hat{m}}\ge\delta$ with $\hat{m}<L$.  To characterize how far ahead of the
true event the model can raise an alarm, we compute the Time-to-Accident (TTA)
as $\mathrm{TTA} = \lambda - \hat{m}$, where $\lambda$ marks the ground-truth frame at which the accident begins, and
we set $\lambda=0$ for sequences without accidents.  Larger TTA values indicate
that the system identifies dangerous situations earlier, thus demonstrating a
higher level of anticipatory ability.  The overall objective is to learn a
reliable risk curve $\mathcal{U}$ while maximizing the advance-detection index
$\hat{m}$ for videos in which an accident occurs.


\subsection{Traffic Video Synthesis}

Figure \ref{Overall} shows our entire model framework. We first start with the generation of synthetic data. This module generates driving videos for data augmentation to tackle the data insufficient.

\begin{figure}[t]
\centering  
\includegraphics[width=0.7\textwidth]{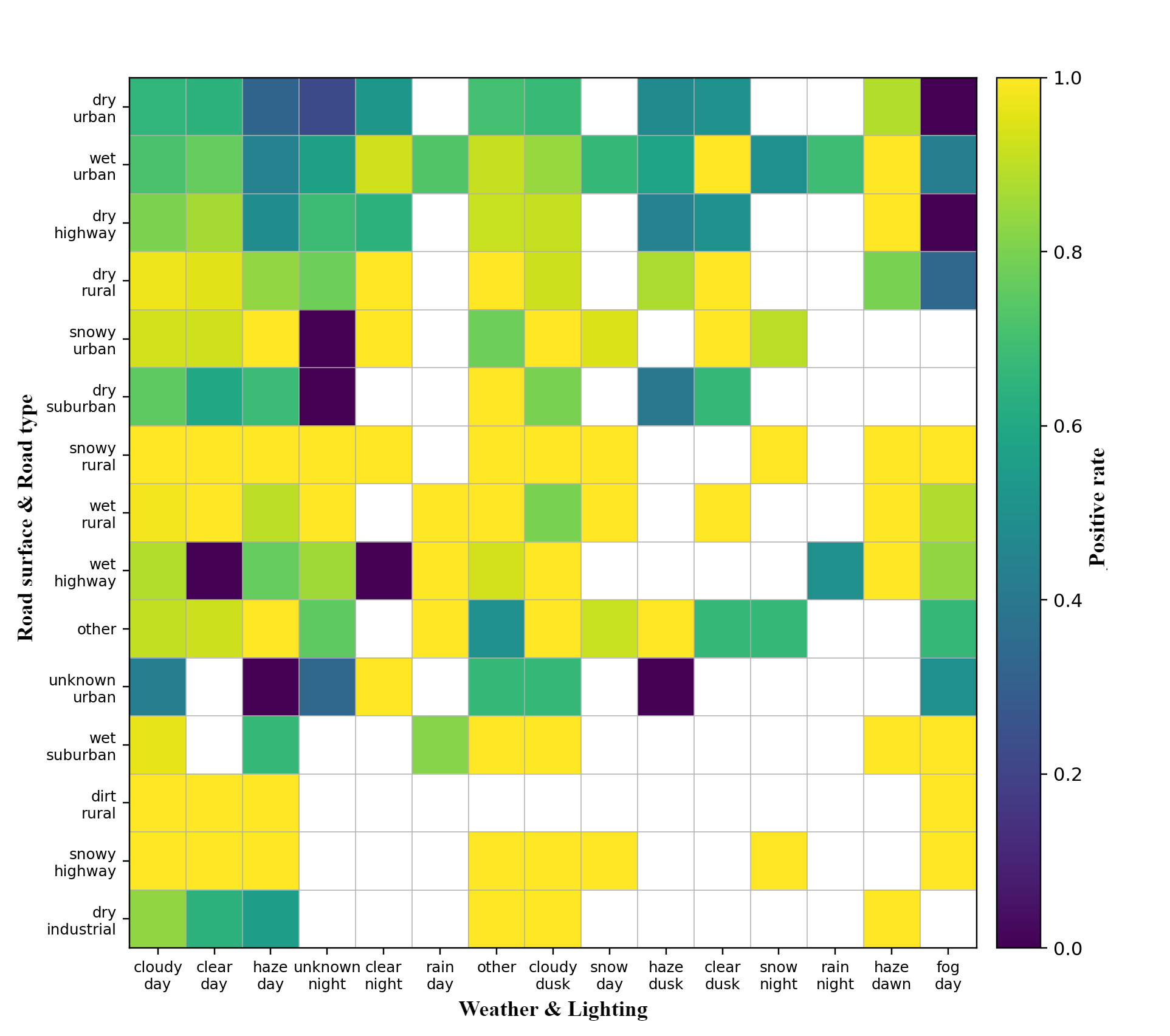}
\caption{Heatmap of positive sample proportions under different environment in MAA dataset. Within the dataset, accidents' occurrence varies substantially across different conditions. Moreover, despite being the largest accident anticipation dataset to date, certain environments still lack sufficient samples.
}
\label{Factors}
\end{figure}

Given the restricted diversity of existing datasets, we propose a controllable traffic video synthesis framework for synthesizing plausible driving videos. The proposed framework leverages environmental distributions extracted from existing data and reconstructs them into novel scenarios, ensuring that the generated samples preserve the statistical properties of the original dataset. Concretely, we first employ Qwen-VL \cite{bai2023qwen} to obtain the underlying environmental distributions (Figure \ref{Factors}). When the testing samples remain fixed, preserving the original distribution of the dataset helps prevent the training process from being contaminated by environmental conditions absent from the dataset. However, for broader real-world applications, it is essential to address the missing environmental feature distributions, enabling the model to learn from more diverse scenarios.
Next, random mapping based on the extracted distributions is applied to controllably determine the environment of new scenarios. 

\begin{figure}[t]
\centering  
\includegraphics[width=0.8\textwidth]{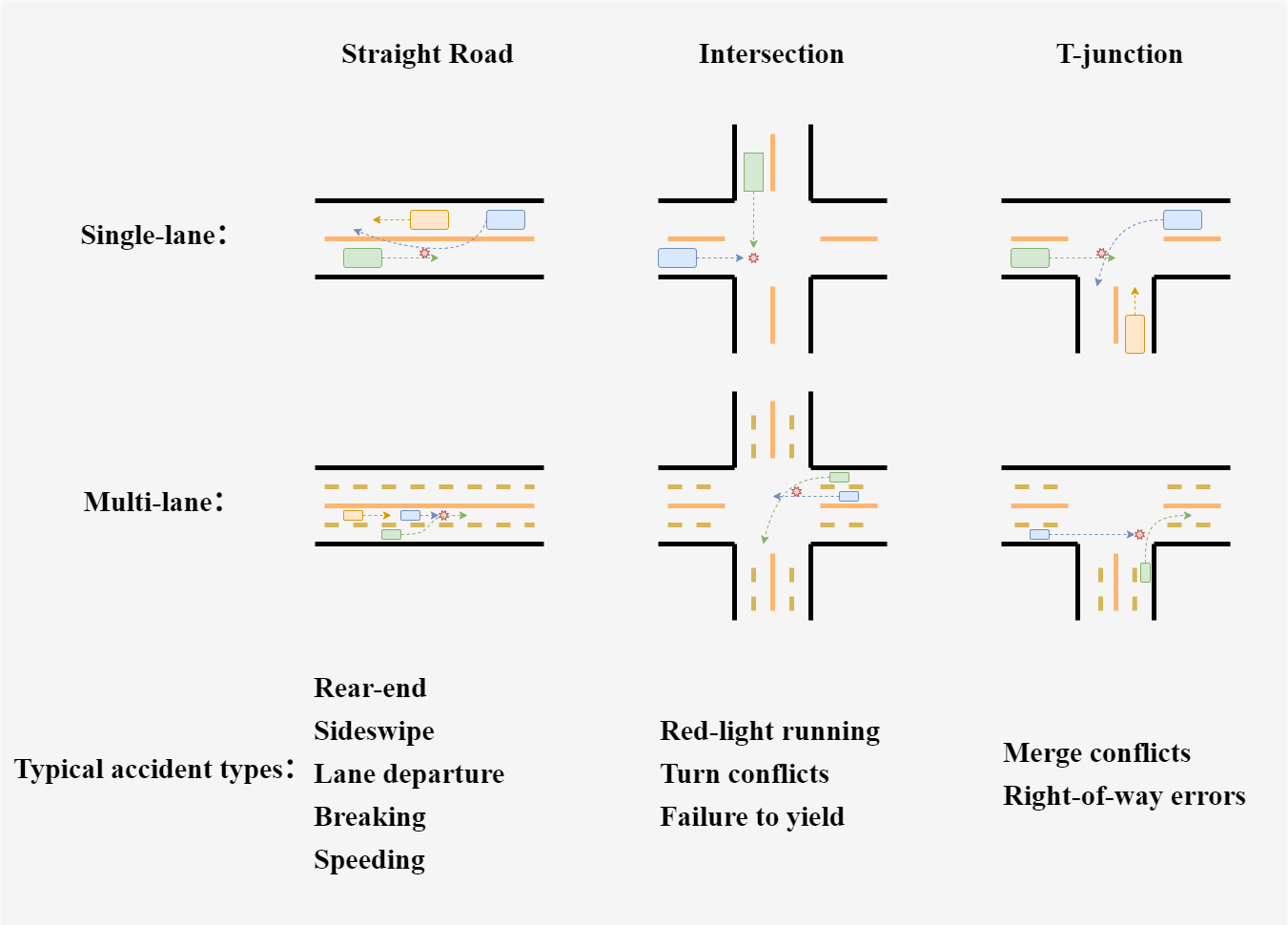}
\caption{The preset maps of accident driving scenarios generation and corresponding accident types. We have designed some preset maps for common scenarios and summarized the corresponding typical accident types to ensure that common accidents can be reconstructed.}
\label{Presetmap}
\end{figure}

We then adopt two different strategies to assign maps for negative and positive videos.  
For positive samples, the inherent complexity of designing accident trajectories necessitates pre-defining both the accident type and its spatial location. To address this challenge, we employ preset maps that simplify trajectory design while ensuring coverage of common accident scenarios. According to the National Motor Vehicle Crash Causation Survey (NMVCCS), more than 70\% of accidents occur on intersections or straight roads, with intersections alone accounting for nearly 40\% \cite{shinar2017accident}. Therefore, as illustrated in Figure \ref{Presetmap}, our preset maps primarily consist of intersections, T-junctions, straight segments, as well as single-lane and multi-lane configurations.  

For negative samples, the map assignment is comparatively straightforward. To increase terrain diversity, we randomly select coordinates from publicly available road data provided by the U.S. National Highway Traffic Safety Administration (NHTSA) and download the corresponding high-definition maps.

Figure~\ref{Overall}b illustrates the key information required for generating driving scenarios. After obtaining the map and defining drivable regions, it is necessary to assign origins and destinations for traffic participants. For participants that do not require fine-grained control, random routes are assigned based on the map and traffic rules, ensuring conflict-free trajectories. Beyond these initial trajectories, each participant’s real-time actions are governed by a driving agent.

Specifically, a directed weighted graph is first constructed based on the road network, denoted as:
\begin{equation}
G = (V, E, w)
\end{equation}
where $V$ represents the set of road nodes,  
$E \subseteq V \times V$ denotes the set of directed edges corresponding to traversable road segments, and
$w$ is the temporal weight function that assigns a travel cost to each edge. 
Each edge $e_{ij} \in E$ is associated with the following attribute:
\begin{equation}
w(e_{ij}) = \frac{L_{ij}}{v_{ij}}
\end{equation}
where $L_{ij}$ denotes the road length and $v_{ij}$ represents the speed limit.

The sets of traversable and non-internal edges are defined as the origin and destination edge sets, denoted by $E_{\mathrm{src}}$ and $E_{\mathrm{dst}}$, respectively.  
For each vehicle $k$, its starting and ending edges are obtained by random sampling:
\begin{equation}
e_s(k) \sim E_{\mathrm{src}}, \qquad
e_t(k) \sim E_{\mathrm{dst}}
\end{equation}

Vehicle departures are modeled as a Poisson process with rate $\lambda$.  
Let $\{N(t)\}_{t\ge 0}$ denote the counting process of vehicle insertions, then
$N(t) \sim \mathrm{Poisson}(\lambda t)$
and the inter-arrival times $\Delta t_k$ follow an exponential distribution:
\begin{equation}
\Delta t_k \sim \mathrm{Exp}(\lambda), \qquad
t(k) = \sum_{i=1}^{k} \Delta t_i
\end{equation}
where $t(k)$ is the departure time of vehicle $k$, $\lambda = \tfrac{p}{b}$, $p$ denotes the expected number of vehicle departures within a time window of length $b$.

For each origin–destination (OD) edge pair $\big(e_{s}(k), e_{t}(k)\big)$,  
the shortest path is computed using the Dijkstra~\cite{wang2011application} algorithm:
\begin{equation}
\pi(k) = \arg\min_{\pi \in \Pi\big(e_{s}(k), e_{t}(k)\big)} 
\sum_{e_{ij} \in \pi} w(e_{ij})
\end{equation}
where $\Pi\big(e_{s}(k), e_{t}(k)\big)$ denotes the set of all feasible paths from the origin to the destination.
The corresponding path length is given by:
\begin{equation}
L(k) = \sum_{e_{ij} \in \pi(k)} L_{ij}
\end{equation}

Given a simulation horizon $T_{\mathrm{sim}}$, the expected number of vehicles is:
\begin{equation}
\mathbb{E}[N] = \lambda T_{\mathrm{sim}}
\end{equation}
and in our implementation, we set $T_{\mathrm{sim}} = 6\,$s and discard the first and last $0.5\,$s when rendering, in order to stabilize the number of visible vehicles and the visual quality.

The entire set of random trips (OD, route, and time for each vehicle) is defined as:
\begin{equation}
T = \left\{
\big(e_{s}(k), e_{t}(k), t(k), \pi(k)\big)
\,\middle|\, 
k = 1, 2, \ldots, N
\right\}
\end{equation}

For any trip $\big(e_{s}(k), e_{t}(k), t(k), \pi(k)\big)$,  
the continuous geometric curve along $\pi(k)$ is formed by concatenating the edge-wise centerlines:
\begin{equation}
r^{(k)}(s) = \mathrm{concat}\big(r_{e_1}(s),\, r_{e_2}(s),\, \ldots \big),
\qquad s \in [0, L(k)]
\end{equation}

According to the edge speed $v_e$, a temporal mapping $t(s)$ is established as:
\begin{equation}
t(s) = t(k) + \int_{0}^{s} \frac{1}{v(\sigma)}\,\mathrm{d}\sigma 
\end{equation}

For each temporal sample $t_n^{(k)}$, the spatial position is obtained by inverting the mapping $t(s)$ to find $s_n$ such that $t(s_n) = t_n^{(k)}$, and evaluating the curve $r^{(k)}(s_n)$. The trajectory is sampled with a fixed time step $\Delta t$ as:
\begin{equation}
X^{(k)} = 
\left\{
\big(x_{n}^{(k)},\, y_{n}^{(k)},\, t_{n}^{(k)}\big)
\right\}_{n=1}^{N_k},
\qquad
t_{n}^{(k)} = t(k) + n \Delta t
\end{equation}
where $N_k = \left\lfloor T^{(k)} / \Delta t \right\rfloor$ and $T^{(k)}$ is the travel time of vehicle $k$.

All trajectories collectively constitute the output trajectory set:
\begin{equation}
X = \Phi(T) = 
\left\{
X^{(1)},\, X^{(2)},\, \ldots,\, X^{(N)}
\right\}
\end{equation}

After obtaining the initial trajectories, UniAD \cite{hu2023planning} is employed as the driving agent of each vehicle to establish a closed-loop control system among trajectories, rendered images, and the agent itself.  
When no surrounding vehicles are present within the ego vehicle’s interaction range, the ego vehicle follows its predefined trajectory $X^{(i)}$.  
When potential interactions with nearby vehicles occur, driving agent infers the motion relationships between the ego and surrounding vehicles based on real-time rendered images.  
It then combines this perception with the target route to control the ego vehicle, yielding an updated driving trajectory denoted as $\hat{X}^{(i)}$.

\begin{figure}[t]
\centering  
\includegraphics[width=0.7\textwidth]{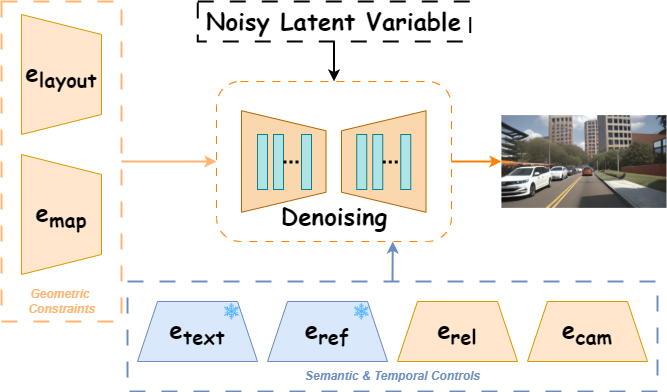}
\caption{The denoising process of scenes' rendering. Geometric constraints use vehicle projection and maps as mandatory constraints, while semantic and temporal controls guide the content of the generated images in terms of semantics and temporal continuity.
}
\label{Denoising}
\end{figure}

For accident-related vehicles and the ego vehicle in positive scenarios, trajectories are deterministically generated to ensure that the accident occurs within the ego vehicle’s dashcam field of view at a specific time step. Therefore, after obtaining $\Pi\big(e_{s}(k), e_{t}(k)\big)$, we feed the textual description of the accident type, the candidate OD pairs of the primary vehicles, and the road network map in XML format into Qwen3-Max \cite{bai2023qwen}. The LLM is instructed to stochastically generate multiple trajectory scenarios corresponding to the same accident type under the following constraints:  
(i) the origins and destinations of all accident-involved vehicles must satisfy the input OD pairs;  
(ii) the trajectories of the accident vehicles must contain at least an intersection point to ensure that a collision occurs;  
(iii) an additional ego vehicle must be present in the scene, and the accident location at the moment of collision must lie within the ego vehicle’s forward field of view, specifically within the $\pm 30^{\circ}$ range of the front-facing camera; and  
(iv) the generated trajectories for all vehicles must be provided with the accident time explicitly annotated.  
All generated samples are then manually inspected, and those that exhibit clearly unreasonable behaviors are discarded.

After generating trajectories for all agents, the environmental distributions extracted from the original dataset are used as prompts to guide a pretrained stable diffusion model trained in nuScenes \cite{caesar2020nuscenes} in rendering the scenes. The denoising process is shown in Figure \ref{Denoising}.
Let the diffusion timestep be denoted as $\tau$, and the noise prediction network as $\epsilon_\theta$. 
The reverse diffusion process is defined as:
\begin{equation}
z_{\tau-1} = f_{\theta}(z_{\tau},\, e_{\mathrm{cond}})
\end{equation}
where the conditional embedding is given by:
\begin{equation}
e_{\mathrm{cond}} = 
\{ e_{\mathrm{text}},\, e_{\mathrm{cam}},\, e_{\mathrm{map}},\, 
e_{\mathrm{layout}},\, e_{\mathrm{ref}},\, e_{\mathrm{rel}} \}
\end{equation}

Here, $e_{\mathrm{text}}$ is obtained using a CLIP text encoder.  
The intrinsic and extrinsic camera parameters $e_{\mathrm{cam}}$ are encoded via Fourier embeddings.  
The BEV grid $e_{\mathrm{map}}$ is extracted using a convolutional encoder with positional encoding.  
All lane and boundary maps, together with vehicle bounding boxes in BEV, are projected onto the image plane of each camera.  
The concatenation of the lane and vehicle projections is then fed into a conditional encoder to produce $e_{\mathrm{layout}}$. The reference embedding $e_{\mathrm{ref}}$ corresponds to a randomly sampled historical frame,  
introduced to maintain temporal semantic consistency.  
Let the current ego pose be $T_{t}$ and the reference pose be $T_{r}$;  
the relative transformation is given by:
\begin{equation}
\Delta T = T_{t} T_{r}^{-1}
\end{equation}
and its Fourier embedding is expressed as:
\begin{equation}
e_{\mathrm{rel}} = \gamma(\Delta T)
\end{equation}
which enforces temporal consistency of the background. To ensure geometric coherence among multi-camera views, a cross-view attention mechanism is introduced:
\begin{equation}
h_i' = \mathrm{Attn}\big(h_i,\, \{h_j\}_{j \neq i}\big)
\end{equation}
where $h_i$ denotes the latent feature corresponding to the $i$-th camera.

During inference, an autoregressive strategy is adopted to enhance temporal stability. For the $t$-th generated frame, the reference embedding is obtained from the previously generated frame $t-1$, and the relative ego-pose embedding is derived from the motion between consecutive poses:
\begin{equation}
e_{\mathrm{ref}}^{(t)} = \mathrm{CLIP}(I_{t-1}), \quad
e_{\mathrm{rel}}^{(t)} = \gamma(T_{t} T_{t-1}^{-1})
\end{equation}
leading to the final reverse diffusion based generation for frame $t$:
\begin{equation}
I_{t} = g_{\theta}\big(
z_{\tau}^{(t)},\,
e_{\mathrm{text}},\,
e_{\mathrm{cam}},\,
e_{\mathrm{map}},\,
e_{\mathrm{layout}},\,
e_{\mathrm{ref}}^{(t)},\,
e_{\mathrm{rel}}^{(t)}
\big)
\end{equation}
where $z_{\tau}^{(t)}$ denotes the noisy latent variable at diffusion step $\tau$ for frame $t$.

\begin{figure}[t]
\centering  
\includegraphics[width=0.8\textwidth]{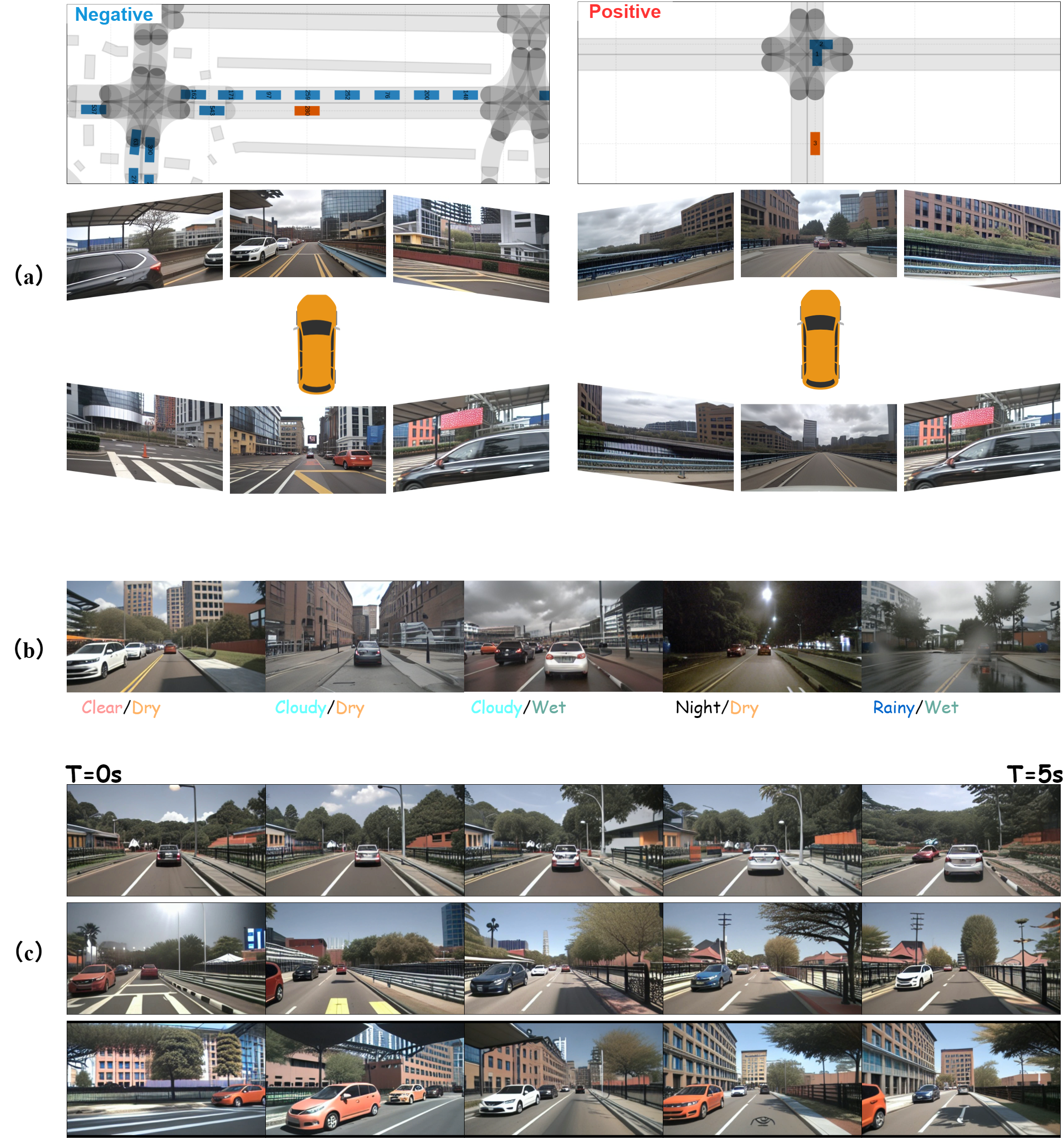}
\caption{Examples of the generated videos. (a) Multi-view videos generated for accident and non-accident scenarios along with corresponding high-definition maps and traffic flows. (b) Video examples of different weather and road conditions. (c) Examples of primary perspective videos in chronological order.}
\label{Videos}
\end{figure}

Figure \ref{Videos} shows some examples of actual synthetic traffic scenes, including HD maps and corresponding multi-view images of both negative and positive scenes, rendering examples in different weather conditions, and some synthetic dashcam videos. 

To evaluate the quality of the synthetic traffic scene videos, we conduct a comparative study using the DAD dataset as a reference and benchmark against another synthetic dataset, EMM-AU \cite{li2025avd2}. The evaluation metrics include Fréchet Inception Distance (FID), Inception Score (IS), Fréchet Video Distance (FVD), Peak Signal-to-Noise Ratio (PSNR), Structural Similarity Index (SSIM), and CLIP similarity \cite{obukhov2020quality,ge2024content,korhonen2012peak,brunet2011mathematical,xu2021videoclip}.

\begin{table*}[t]

\caption{Visual quality comparison of synthetic videos. \textbf{Bold} numbers indicate superior performance for the corresponding metric.}

\resizebox{\textwidth}{!}{
\setlength\tabcolsep{15pt}
\begin{tabular}{ccccccc}
\hline
Dataset  & FID    & IS & FVD   & PSNR  & SSIM & CLIP \\ \hline
EMM-AU \cite{li2025avd2} & 175.58 & \textbf{2.71}  & \textbf{30.87} & 9.38  & 0.39 & 0.84 \\
Ours & \textbf{91.28}  & 2.17  & 34.46 & \textbf{10.18} & \textbf{0.40}  & \textbf{0.84} \\\hline
\end{tabular}}
\label{FID}
\end{table*}

The experimental results in Table~\ref{FID} demonstrate that, compared with EMM-AU, our generated videos are substantially closer to the real distribution, achieving superior fidelity and clarity. However, in terms of temporal consistency and discriminability, our method is slightly weaker than EMM-AU. Moreover, under the same benchmarks, both approaches exhibit comparable levels of semantic alignment. From a visual standpoint, our generated videos display certain imperfections in lane line consistency and distortions in the continuous shapes of vehicles and buildings. Nevertheless, compared with EMM-AU, since our generative model is employed solely for rendering, the synthesized videos avoid unrealistic vehicle dynamics and remain kinematically closer to real-world behavior.

\subsection{Accident Anticipation Framework}

\subsubsection{Video Preprocessing}

In the initial stage of video processing, as shown in Figure \ref{Preprocessing}, it is essential to rapidly and accurately detect vehicles, pedestrians, and other traffic-related entities in each frame. We first apply YOLOv8 \cite{sohan2024review}, a lightweight and high-performance object detection model widely adopted in industrial applications, to perform object detection on the input videos. Subsequently, the number of detected objects per frame is capped at 19, and the ByteTracker \cite{zhang2022bytetrack} algorithm is utilized to assign a consistent ID across frames for each target. Visual features are extracted from both the entire frame and the detected objects using a VGG16 backbone \cite{simonyan2014very}, while depth information for every pixel is estimated with the ZOE Depth model \cite{bhat2023zoedepth}. In addition, Qwen-VL \cite{bai2023qwen} is employed to generate behavioral descriptions of each target and a scene-level summary, which are then matched with their corresponding visual IDs. Finally, textual information is encoded with BERT \cite{aftan2023survey} and aligned with the visual feature dimensions, thereby completing the preprocessing stage. The data structure after the above data processing is as follows:

\begin{figure}[t]
\centering  
\includegraphics[width=0.8\textwidth]{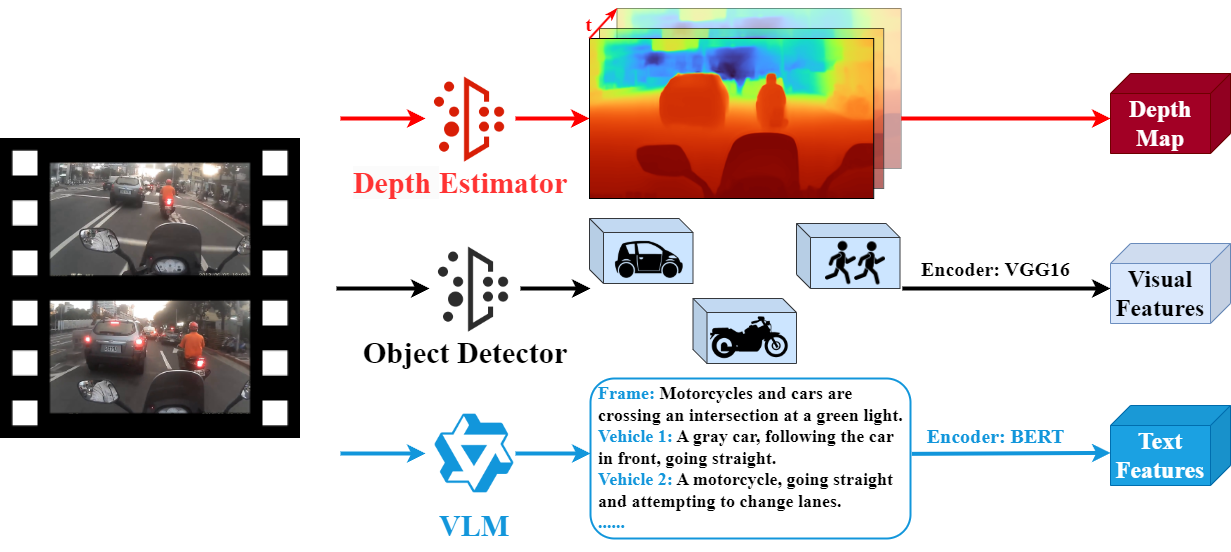}
\caption{Video preprocessing framework. We preprocess each video to obtain the key info required for inference, including object- and frame-level visual features, semantic features, and geometric cues.}
\label{Preprocessing}
\end{figure} 

\begin{equation}
\begin{aligned}
&F_{vis} \in \mathbb{R}^{B \times T \times N \times F} \\
&Det \in \mathbb{R}^{B \times T \times (N-1) \times F} \\
&F_{text} \in \mathbb{R}^{B \times T \times N \times F} \\
&Depth \in \mathbb{R}^{B \times T \times W \times H}
\end{aligned}
\end{equation}

where $F_{vis}$, $Det$, $F_{text}$, and $Depth$ refer to the visual feature
representation, the object detection output, the text-derived feature tensor, 
and the depth information.  
The symbols $B$, $T$, $N$, and $F$ indicate the batch count, the temporal length 
of the sequence, the total number of aggregated features (covering both frame-level 
and object-level cues), and the dimensionality of each feature vector.  
For the depth modality $Depth$, the parameters $W$ and $H$ specify the spatial 
resolution of each frame.

\subsubsection{Graph Construction with Geometric and Semantic Guided Connectivity}


To capture spatial interactions among traffic participants, we model each frame as a graph where objects are nodes and pairwise relations form edges. The adjacency matrix is adaptively learned and further refined by both geometric and semantic cues.

Given detected objects per frame $O = N-1$ , we parameterize the adjacency matrix with low-rank factorization:
\begin{equation}
A = \phi_{\textit{softmax}}(UV) \in \mathbb{R}^{O \times O}, \quad 
\tilde{A} = D^{-\frac{1}{2}} (A + I) D^{-\frac{1}{2}}
\label{eq:adj}
\end{equation}
where $U \in \mathbb{R}^{O \times O}$ and $V \in \mathbb{R}^{O \times O}$ are learnable embeddings, $I$ is the identity matrix, and $D$ is the diagonal degree matrix. The role of this adaptive adjacency matrix is to dynamically model the relationship between objects in a learnable way, and to ensure stable and reasonable feature aggregation by determining which objects will interact with each other and the strength of the interaction.

Traffic participants located near one another exhibit an increased risk of potential crashes, particularly when their speed differences are large. In contrast, entities separating from each other are less likely to be engaged in hazardous interactions. Therefore, to enrich connectivity with physical constraints, we compute the 3D geometric distances and relative velocities between objects.  
For each video, let $c_{t,i}$ denotes the 2D center of object $i$ at frame $t$, and $z_{t,i}$ denotes its depth from $Depth$. The combined distance is defined as:
\begin{equation}
d_{t,ij} = s^2 \lVert c_{t,i} - c_{t,j} \rVert_2^2 + \lvert z_{t,i} - z_{t,j} \rvert^2
\label{eq:dist}
\end{equation}
where $s$ is a normalization constant. The relative velocity is estimated by frame-wise differences:
\begin{equation}
v_{t,ij} = d_{t,ij} - d_{t-1,ij}
\label{eq:vel}
\end{equation}

After normalization, geometry-driven edge weights are given by:
\begin{equation}
W_{t,ij}^{geo} = \alpha \exp(-\bar{d}_{t,ij}) + (1 - \alpha)\,\bar{v}_{t,ij}, 
\quad \alpha = \frac{a}{a+1}
\label{eq:geo}
\end{equation}
with the initial balance parameter $a=1$.

In addition to visual-spatial relationships, we explicitly model semantic object interactions in the construction of the graph structure, thereby simultaneously considering the spatial and semantic dependencies between objects. This significantly improves the ability to model long-term dependencies and causal reasoning, making the model's understanding of complex scenes more comprehensive and accurate. For semantics, object-level textual embeddings aligned with IDs are normalized, and cosine similarities are converted into text-driven edge weights:
\begin{equation}
W_{t,ij}^{text} = 
\frac{\exp\big(\langle \hat{x}_{t,i}^{text}, \hat{x}_{t,j}^{text} \rangle / \tau_{text}\big)}
{\sum_{j'} \exp\big(\langle \hat{x}_{t,i}^{text}, \hat{x}_{t,j'}^{text} \rangle / \tau_{text}\big)}
\label{eq:text}
\end{equation}
where $\hat{x}_{t,i}^{text}$ denotes normalized text features and $\tau_{text}$ is a temperature parameter.

The overall edge weight matrix combines geometry- and text-driven terms via a learnable gate:
\begin{equation}
W_t = (1 - \lambda) W_t^{geo} + \lambda W_t^{text}, 
\quad \lambda = \sigma(\beta)
\label{eq:fusion}
\end{equation}
where $W_t \in \mathbb{R}^{O \times O}$ encodes both geometry-aware and text-guided connectivity, and $\lambda$ is dynamically controlled by $\beta$, which allows our model to learn the distribution of semantic and geometric weights in different scenarios during training.

Once the structural graph is constructed, we perform multi-granularity multimodal fusion to ensure semantic consistency across modalities. Specifically, at the object level, we align each detected entity’s visual embedding with its corresponding textual embedding and apply a gating mechanism to adaptively emphasize the more reliable modality. This design allows the model to preserve fine-grained semantic cues while remaining robust to noise or missing information in either modality. At the frame level, we further integrate global visual scene embeddings with sentence-level textual descriptions to provide a coherent context that guides object interactions. Such hierarchical fusion guarantees that both local interactions and global scene semantics are consistently represented.

For each object $i$ at frame $t$, let $x_{t,i}^{vis} \in \mathbb{R}^F$ denote its visual embedding extracted from $F_{vis}$, and $\hat{x}_{t,i}^{text} \in \mathbb{R}^F$ denote the aligned textual embedding. We employ a gating mechanism to adaptively combine the two representations:
\begin{equation}
x_{t,i}^{fuse} = g\!\left([x_{t,i}^{vis}; \hat{x}_{t,i}^{text}]\right) \odot x_{t,i}^{vis} 
+ \left(1 - g(\cdot)\right) \odot \hat{x}_{t,i}^{text}
\label{eq:obj_fusion}
\end{equation}
where,
\begin{equation}
g(\cdot) = \sigma\!\left(W_g [x_{t,i}^{vis}; \hat{x}_{t,i}^{text}] + b_g\right)
\end{equation}
$\sigma(\cdot)$ is the sigmoid activation, and $[;]$ denotes feature concatenation.

Similarly, we fuse frame-level features. Let $f_t^{vis}$ be the global visual embedding from $F_{vis}$, and $\hat{f}_t^{text}$ the sentence-level embedding from $F_{text}$. Their fusion is defined as:
\begin{equation}
f_t^{fuse} = g\!\left([f_t^{vis}; \hat{f}_t^{text}]\right) \odot f_t^{vis} 
+ \left(1 - g(\cdot)\right) \odot \hat{f}_t^{text}
\label{eq:frame_fusion}
\end{equation}

\subsubsection{Temporal Modeling}
To jointly capture spatial and temporal dependencies, we propose a dynamic weighted graph convolutional network followed by multi-scale temporal encoders.

Given fused object-level features $X_t^{fuse} = [x_{t,1}^{fuse}, \ldots, x_{t,O}^{fuse}]$, the GCN layer updates node embeddings as:
\begin{equation}
H_t^{(l+1)} = \sigma \!\left( W_t \odot \tilde{A}_t H_t^{(l)} \Psi^{(l)} \right)
\label{eq:gcn}
\end{equation}
where $H_t^{(0)} = X_t^{fuse}$, $\tilde{A}_t$ is the normalized adjacency, and $\Psi^{(l)}$ are learnable weights. After $L = 2$ layers, node embeddings are pooled:
\begin{equation}
u_t = \frac{1}{O} \sum_{i=1}^O H_{t,i}^{(L)}
\label{eq:pool}
\end{equation}

The pooled representation $u_t$ is concatenated with $f_t^{fuse}$ to form $z_t$. To enlarge temporal receptive fields, we apply temporal convolutional network:
\begin{equation}
\tilde{z}_{1:T} = \text{TCN}(z_{1:T})
\end{equation}
followed by a GRU for sequential modeling:
\begin{equation}
h_{1:T} = \text{GRU}(\tilde{z}_{1:T})
\end{equation}

Frame-level predictions are computed as:
\begin{equation}
\ell_t = W_2 \,\phi(W_1 h_t + b_1) + b_2, 
\quad p_t = \text{softmax}(\ell_t)
\label{eq:pred}
\end{equation}
where $\ell_t \in \mathbb{R}^2$ are accident anticipation logits.

\subsubsection{Loss Function}
The model is supervised with three complementary objectives.

\paragraph{(i) Frame-level anticipation loss.}
Positive samples are reweighted according to their distance from the time-of-accident $\tau$, which encourages models to make predictions as early as possible and with greater confidence closer to the moment of an accident:
\begin{equation}
w_t^+ = \exp\!\left(-\max(0, \tau - t - 1/fps)\right)
\end{equation}
and the frame-level anticipation loss is
\begin{equation}
L_1 = \frac{1}{BT}\sum_{b=1}^B \sum_{t=1}^T \Big( y\, w_t^+ \,\text{CE}(\ell_t,y) 
+ (1-y)\,\text{CE}(\ell_t,y) \Big)
\end{equation}

\paragraph{(ii) Video-level pooling loss.}
From the perspective of the entire positive scene video, at least one frame in the first $T_{pool}$ frames should be sufficient to determine the occurrence of an accident. Therefore, video-level classification loss is applied to improve the accuracy of the model's accident prediction:
\begin{equation}
L_2 = \frac{1}{B} \sum_{b=1}^B \text{CE}\!\left(\max_{t \leq T_{pool}} \ell_t, y\right)
\end{equation}

\paragraph{(iii) Semantic alignment loss.}
To enforce cross-modal semantic consistency, we impose a contrastive constraint between visual and textual representations. For each time step \(t\), we take the global visual embedding \(f_t^{\mathrm{vis}}\) and the corresponding sentence-level textual embedding \(\hat{f}_t^{\mathrm{text}}\), and apply a temperature-scaled, bidirectional Information Noise-Contrastive Estimation (InfoNCE) objective to pull the cross-modal positive pair at the same time step closer while pushing other samples/time steps in the mini-batch apart as negatives:

\begin{equation}
\begin{aligned}
&L_{v\to t}(t)
=
-\log
\frac{
\exp\!\big(\langle \bar f_{t}^{\mathrm{vis}},\, \bar f_{t}^{\mathrm{text}} \rangle / \tau_c\big)
}{
\sum_{(b',\,t')\in \mathcal{B}}
\mathbf{m}_{(t),(b',t')}
\exp\!\big(\langle \bar f_{t}^{\mathrm{vis}},\, \bar f_{t'}^{\mathrm{text},(b')} \rangle / \tau_c\big)
}\\
&L_{t\to v}(t)
=
-\log
\frac{
\exp\!\big(\langle \bar f_{t}^{\mathrm{text}},\, \bar f_{t}^{\mathrm{vis}} \rangle / \tau_c\big)
}{
\sum_{(b',\,t')\in \mathcal{B}}
\mathbf{m}_{(t),(b',t')}
\exp\!\big(\langle \bar f_{t}^{\mathrm{text}},\, \bar f_{t'}^{\mathrm{vis},(b')} \rangle / \tau_c\big)
}\\
&L_3 = \frac{1}{|T|} \sum_{t \in T} \frac{1}{2}\Big(L_{v \to t}(t) + L_{t \to v}(t)\Big)
\end{aligned}
\end{equation}

For numerical stability, we perform \(\ell_2\) normalization before projection and use a lightweight MLP as the projection head; we also adopt a temporal-neighborhood mask to avoid treating adjacent frames from the same video as negatives. Serving as an orthogonal regularizer to the main task, this loss strengthens cross-modal alignment, encourages the object/frame-level fusion gates to learn more reliable evidence weighting, and stabilizes the text-driven edge weights in the graph, thereby improving the modeling of long-range dependencies and causal cues.

\paragraph{(iv) Final objective.}
The overall training loss is:
\begin{equation}
L = L_1 + \gamma L_2 + L_3, 
\quad \gamma=T
\end{equation}
where, $L_1$ is a frame-level loss, and $L_2$ is a video-level loss. The frame number $T$ is used as an amplification factor to prevent $L_2$ from being overwhelmed by $L_1$. $L_3$ serves as an auxiliary loss to stabilize cross-modal alignment, achieving the complementary advantages of frame-level fine supervision, video-level early anticipation constraints, and cross-modal consistency.

\section{Experiments}
\label{sec:Experiments}


We assess the effectiveness of our pipeline on four real-world datasets: the Dashcam Accidents Dataset (DAD) \cite{chan2017anticipating}, the AnAn Accident Detection (A3D) dataset \cite{yao2019unsupervised}, The Car Crash Dataset (CCD) \cite{BaoMM2020} and the Multi-source Accident Anticipation (MAA) dataset introduced in this work. The detailed information of the existing datasets is shown in Table \ref{datasets}.

The MAA dataset is introduced as a benchmark dedicated to accident anticipation, constructed by integrating multiple publicly available video resources such as D²-City, BDD100K, and DOTA \cite{yao2022dota,che2019d,yu2020bdd100k}. MAA consists of 6,000 clips, of which two-thirds are accident-positive and the rest are negative samples. To ensure fairness, we re-processed all clips to maintain consistent resolution, frame rate, and duration, and re-annotated the accident start time within a specific frame interval.
The dataset is partitioned into 4,500 samples for training and 1,500 for testing. A detailed comparison between MAA and existing datasets is provided in Table~\ref{datasets}. Compared to other datasets, MAA not only contains more videos but also offers greater coverage. Because MAA draws from multiple sources, its scenes span diverse geographic regions, including Asia, the Americas, and internet videos (while DAD is limited to a single region). Furthermore, MAA encompasses a wider range of climate and traffic conditions, along with richer annotations, providing a more comprehensive resource for accident prediction research. We have deployed several representative baseline models on MAA to provide performance references for future research.

\begin{table*}[t]
    \centering
    \caption{The detailed information of MAA and other existing datasets. \textit{mToA} represents the mean start time of accident.}
    \setlength\tabcolsep{10pt}
     \resizebox{1.0\linewidth}{!}{
    \begin{tabular}{ccccccc}
    \hline
        Dataset & Videos & Positives & Negative & Total Length (h)& FPS  &mToA \\ \hline
        DAD \cite{chan2017anticipating}& 1750 & 620 & 1130 &2.43& 20  &4.50  \\
        DADA-2000 \cite{fang2019dada} &2000 &2000 &- &6.09&30&-\\
        CCD \cite{BaoMM2020}& 4500 & 1500& 3000  &6.25& 10  &3.82 \\
        CADP \cite{shah2018cadp}&1416&1416&-&5.2&28&3.69\\
        A3D \cite{yao2019unsupervised}& 1500 & 1500 & - &2.08& 20   &3.37\\
        MAA (Ours) & 6000 & 4000 & 2000 &8.33& 10  &3.16 \\
    \hline
    \end{tabular}
    }
    
    \label{datasets}
\end{table*}

\begin{table*}[t]
\centering
\caption{{Comparison of models on DAD, A3D, CCD, MAA and Synthetic DAD (sDAD). \textbf{Bold} and \underline{underlined} values represent the best and second-best performance. Instances where values are not available are marked with a dash (``-'').}}
\setlength\tabcolsep{2pt}
 \resizebox{\linewidth}{!}{
\begin{tabular}{lcccccccccc}
\hline \specialrule{0em}{1pt}{1pt}
\multirow{2}[2]{*}{Model} & \multicolumn{2}{c}{DAD}& \multicolumn{2}{c}{A3D}& \multicolumn{2}{c}{CCD} & \multicolumn{2}{c}{MAA} & \multicolumn{2}{c}{sDAD} \\
\cmidrule(lr){2-3} \cmidrule(lr){4-5} \cmidrule(lr){6-7} \cmidrule(lr){8-9} \cmidrule(lr){10-11}
 & AP (\%)↑ & mTTA (s)& AP (\%) & mTTA (s) & AP (\%) & mTTA (s)& AP (\%) & mTTA (s)& AP (\%) & mTTA (s) \\
\hline
DSA \cite{chan2017anticipating}& 48.1 & 1.34  & 92.3 & 2.95 & 98.7 & 3.08& - & - & - & - \\
VideoMAE \cite{tong2022videomae} & 48.1 & 3.93  & - & - & - & -& \underline{76.5} & 1.90 & 55.1 & \textbf{4.02} \\
AdaLEA \cite{suzuki2018anticipating}& 52.3 & 3.43  & 92.9 & 3.16 & 99.2 & 3.45& -  & -& - & - \\
UString \cite{BaoMM2020}& 53.7 & 3.53  & 93.2 & \textbf{3.24} & 99.5 & \underline{3.74}& 69.9 & 1.61 & 58.2 & 3.10\\
DSTA \cite{karim2022dynamic}& 56.1 & 3.66  & 93.5 & 2.87 & 99.3 & 3.87\footnotemark[1] & 74.5 &  1.85  & 76.5 & 2.21 \\
GSC \cite{wang2023gsc}& 60.4 & 2.55  & 94.9 & 2.62 & 99.4 & 3.68& - & - & - & - \\

AccNet \cite{liao2024real} & 60.8 & 3.58  & \underline{94.9} & 2.62 & 99.5 & \textbf{3.78}& 73.1 & 1.69 & 73.8 & 2.55 \\
VideoCLIP \cite{xu2021videoclip} & 68.5 & 3.75  & - & - & - & -& 74.7 & 1.72 & 72.4 & 3.73 \\
THAT-NET \cite{liu2023net}& \underline{77.8} & 3.64 & - & - & \underline{99.5} & 4.59\footnotemark[1]& - & - & - & -
 \\ 
 ViViT-FE \cite{arnab2021vivit} & 78.8 &\underline{4.03} &-&- & - & -&75.8&\underline{1.97}& \underline{82.1} & \underline{3.86}\\
\hline
\textbf{Ours} & \textbf{85.8} & \textbf{4.11}  & \textbf{95.5} & \underline{3.20} & \textbf{99.7} & 3.68& \textbf{80.8} & \textbf{1.99} & \textbf{90.1} & 3.71 \\
\hline
\end{tabular}
}

\label{table:balance}
\end{table*}

\subsection{Evaluation Results}

Table \ref{table:balance} summarizes comparative results across DAD, A3D, CCD, and MAA using Average Precision (AP) and Mean Time-to-Accident (mTTA) as the main evaluation indicators. On the DAD dataset, our method surpasses existing approaches with gains of 10.3\% in AP and 12.3\% in mTTA. For A3D and CCD, where AP values are close to saturation, our model still shows noticeable improvements.

For the MAA benchmark, we replicate multiple representative baselines, including advanced accident anticipation models based on RNNs and graph neural networks, as well as temporal reasoning frameworks based on Transformers, self-supervised learning, and multimodal models. Evaluation on MAA shows our method outperforming the runner-up by 6.5\% AP and 1\% mTTA, validating the advantage of the semantic–geometric graph design and positioning it as a new benchmark for accident anticipation.

\footnotetext[1]{As shown in Table \ref{datasets}, the mToA of the CCD dataset is 3.82, so results larger than this value cannot be compared on the same scale.}

We further investigate the impact of integrating synthetic data into the DAD benchmark. When 40\% of the training set is expanded with generated videos, most baselines show certain gains. For models with advanced performance, this improvement ranges from 4.2\% to 5.0\%. Additional trials include supplementing the dataset with 40\% real traffic clips and, alternatively, substituting 40\% of the original training data with generated samples. The reported results are AP = 90.9 / mTTA = 4.02 for the former and AP = 83.8 / mTTA = 3.40 for the latter. These findings indicate that, while synthetic augmentation contributes positively, its effectiveness remains inferior to augmenting with genuine traffic footage. Besides, when generated data directly replaces part of the original training set, performance deteriorates, indicating the presence of a measurable domain gap, which is expected when synthetic samples replace real data. Based on the experimental results, the performance difference between the two data sources ranges from 0.9\% to 2.3\%, with this gap further diminishing as the dataset size increases. Moreover, the performance discrepancy between synthetic and real data is substantially smaller than the performance gains achieved through the use of synthetic data. It is worth noting that when additional data is incorporated into the training set, the mTTA metric shows a moderate decline. We observe that the ratio of mTTA to mToA in the DAD dataset reaches 91.3\%, whereas in the larger-scale MAA dataset this ratio drops to 63.0\%. This result aligns with expectations: the expansion of data volume mitigates overfitting observed in smaller datasets, requiring the model to rely on richer contextual information for accident prediction rather than making rapid forecasts solely based on recurring visual patterns.

\begin{figure}[t]
\centering  
\includegraphics[width=0.7\textwidth]{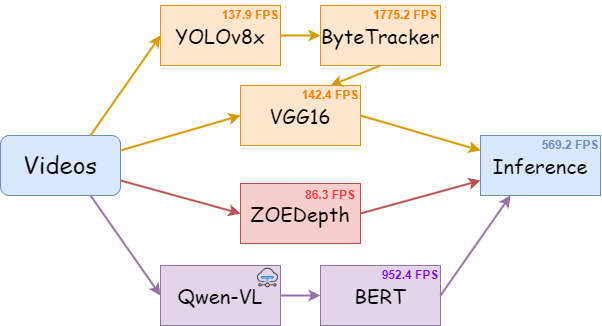}
\caption{The computational efficiency of the key modules within the overall framework. The efficiency of each module is quantified by the average number of video frames it can process per second.}
\label{FPS}
\end{figure}

Considering practical deployment scenarios, we analyze the computational efficiency of the proposed framework. Under the experimental settings, the runtime of each module is shown in Figure \ref{FPS}. Given that Qwen-VL contains approximately 3.4B parameters, on-device deployment is generally infeasible in real-world applications. Therefore, in our efficiency analysis, we assume an edge–cloud collaborative architecture, in which Qwen-VL is deployed in the cloud to accommodate its substantial computational demands. Moreover, because scene semantics evolve slowly across consecutive frames, performing VLM-based scene analysis once every 5 or 10 frames (i.e., every 0.1–0.2 seconds) is sufficient.

In the ideal steady-state pipeline, the throughput is determined by the slowest module (the depth estimation module), yielding a pipeline frame rate of
$FPS_{\text{pipeline}} = 86.3$.
The per-frame output latency corresponds to the cumulative time of the longest sequence of serial operations, which is approximately $16.6\,\mathrm{ms}$. Overall, the depth estimation and VLM modules constitute the primary bottlenecks in the computational pipeline; however, these limitations can be effectively mitigated through cloud-based execution and do not pose significant obstacles to real-world deployment.

\subsection{Ablation Studies}

Table~\ref{Ablation} summarizes the ablation outcomes on the DAD dataset, highlighting how individual components such as the input representations, the GRU unit, the temporal convolutional module and the GCN layers contribute to overall performance. The results show that removing or replacing critical components consistently degrades inference accuracy. Among these modules, visual information, GCN, and GRU contribute most significantly to performance, corresponding to primary information input, scene interaction modeling, and temporal reasoning, respectively. The optimal number of GCN layers is found to be 2, while gated fusion provides the most effective approach for information integration.

\begin{table*}[htbp]
\centering
\caption{Ablation studies on DAD dataset. AAM, TCN represent adaptive adjacency matrix and temporal convolutional network.}
\setlength\tabcolsep{3pt}
\resizebox{\textwidth}{!}{
\begin{tabular}{cccccccccccccc} 
\hline \specialrule{0em}{1pt}{1pt}
\multirow{2}[2]{*}{Model} & \multicolumn{3}{c}{Input}& \multicolumn{3}{c}{GCN layers}& \multicolumn{2}{c}{Fuse} & \multirow{2}[2]{*}{AAM}& \multirow{2}[2]{*}{TCN}& \multirow{2}[2]{*}{GRU} & \multicolumn{2}{c}{Input}\\
\cmidrule(lr){2-4} \cmidrule(lr){5-7} \cmidrule(lr){8-9} \cmidrule(lr){13-14}
 & Vision & Frame Text& Object Text & \multicolumn{3}{c}{1 \quad 2 \quad 3} & Gated & Concat& & & &AP\%&mTTA(s) \\
\hline
A &- &\checkmark &\checkmark &\multicolumn{3}{c}{- \quad \checkmark \quad -} &- & -&\checkmark &\checkmark &\checkmark&44.3&3.12\\

B &\checkmark &- &\checkmark &\multicolumn{3}{c}{- \quad \checkmark \quad -} &\checkmark & -&\checkmark &\checkmark &\checkmark&84.4&3.51\\

C &\checkmark &\checkmark &- &\multicolumn{3}{c}{- \quad \checkmark \quad -} &\checkmark & -&\checkmark &\checkmark &\checkmark&77.4&3.44\\

D &\checkmark &- &- &\multicolumn{3}{c}{- \quad \checkmark \quad -} &- & -&\checkmark &\checkmark &\checkmark&76.1&3.72\\

E &\checkmark &\checkmark &\checkmark &\multicolumn{3}{c}{\checkmark \quad - \quad -} &\checkmark &- &\checkmark &\checkmark &\checkmark&83.2&3.68\\

F &\checkmark &\checkmark &\checkmark &\multicolumn{3}{c}{- \quad  - \quad \checkmark} &\checkmark &- &\checkmark &\checkmark &\checkmark&84.3&3.61\\

G &\checkmark &\checkmark &\checkmark &\multicolumn{3}{c}{- \quad \checkmark \quad -} &- &\checkmark &\checkmark &\checkmark &\checkmark&81.9&3.98\\

H &\checkmark &\checkmark &\checkmark &\multicolumn{3}{c}{- \quad \checkmark \quad -} &\checkmark &- &- &\checkmark &\checkmark&77.7&3.51\\

I &\checkmark &\checkmark &\checkmark &\multicolumn{3}{c}{- \quad \checkmark \quad -} &\checkmark &- &\checkmark &- &\checkmark&78.4&3.85\\

J &\checkmark &\checkmark &\checkmark &\multicolumn{3}{c}{- \quad \checkmark \quad -} &\checkmark &- &\checkmark &\checkmark &-&50.1&3.55\\

K &\checkmark &\checkmark &\checkmark &\multicolumn{3}{c}{- \quad - \quad -} &\checkmark &- &- &\checkmark &\checkmark&46.9&3.90\\
\hline
Final &\checkmark &\checkmark &\checkmark &\multicolumn{3}{c}{- \quad \checkmark \quad -} &\checkmark &- &\checkmark &\checkmark &\checkmark&85.8&4.11\\
\hline
\end{tabular}
}

\label{Ablation}
\end{table*}

\begin{figure}[t]
\centering  
\includegraphics[width=0.9\textwidth]{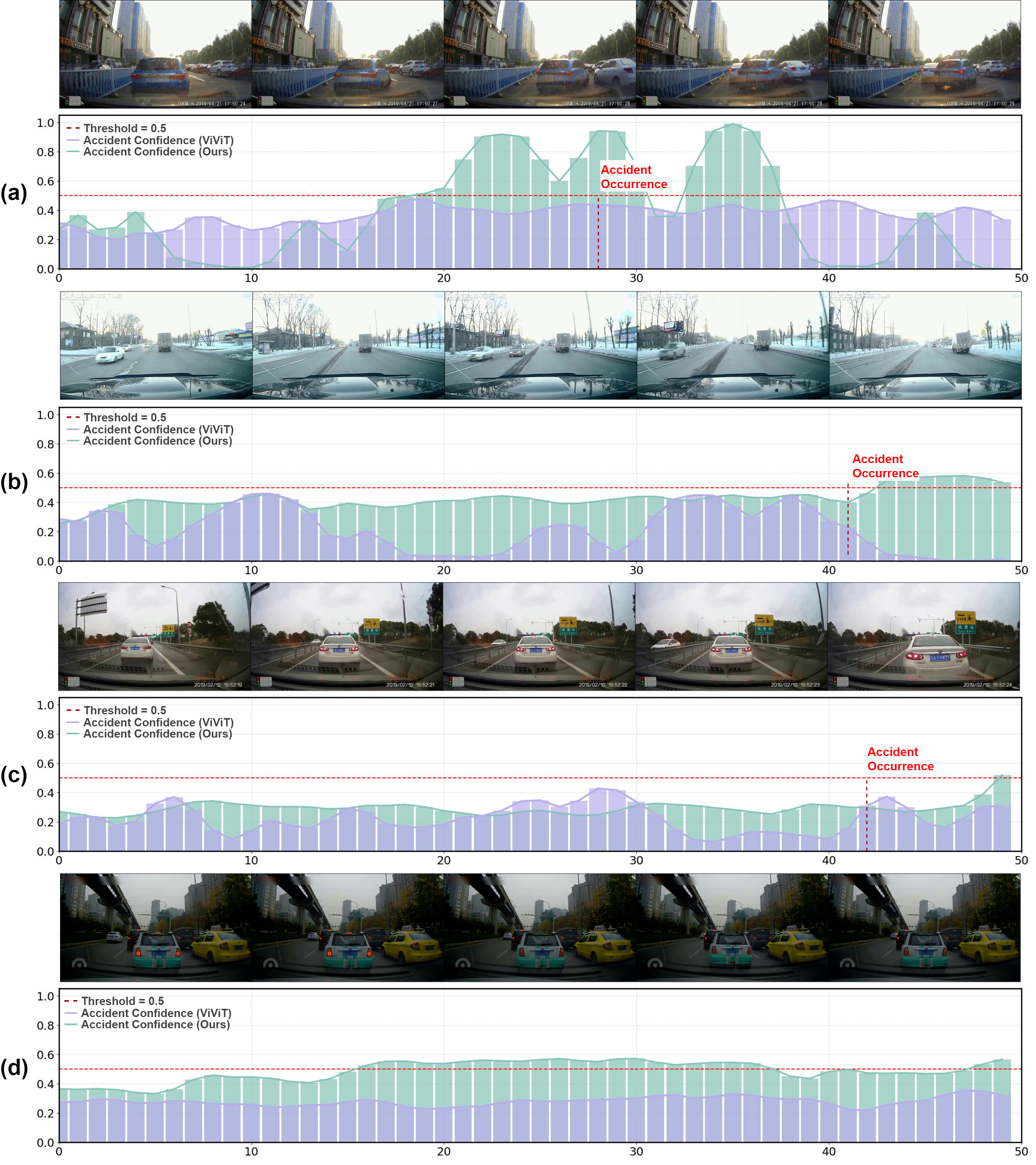}
\caption{Visualization of model performance in some scenes. (a) Successful anticipation of positive scene. (b-c): Failed anticipation of positive scene. (d) Failed anticipation of negative scene.
}
\label{Visualization}
\end{figure}

\subsection{Qualitative Analysis}
To further evaluate the effectiveness of this pipeline, we visualize several traffic scenes from the MAA dataset and compare its performance against the second-best baseline. Figure \ref{Visualization} (a–d) illustrates four challenging scenarios, where Figure \ref{Visualization} (a–c) are positive accident scenes and Figure \ref{Visualization} (d) is a negative case.  

In Figure \ref{Visualization} (a), the blue car in front initiates a rightward lane change while a white car drives straight at high speed along the right lane, ultimately colliding with the blue car without deceleration. From a geometric perspective, the relative speed is high and the distance between the two vehicles is short. From a semantic perspective, a yielding conflict arises between lane changing and straight-line driving. Our model rapidly increases accident confidence and correctly anticipates the accident, with confidence declining after the white car exits the view. By contrast, the baseline model, which relies more heavily on visual features, fails to predict the accident due to the white car’s brief appearance in the camera’s field of view.  

In Figure \ref{Visualization} (b), a heavy truck traveling near the right lane edge runs over a raised roadside curb with its front-right wheel, causing a minor accident. Since no vehicle-to-vehicle interaction occurs and the curb feature is visually obscure due to snow coverage, neither our model nor the baseline successfully anticipates the accident beforehand. Both models only trigger warnings after detecting changes in the truck’s pose.  

Scene Figure \ref{Visualization} (c) depicts a rear-end collision while the ego vehicle is waiting at a red light. Because the colliding vehicle is not visible, both models lack access to critical cues and fail to predict the accident in advance. Notably, in both Figure \ref{Visualization} (b) and Figure \ref{Visualization} (c), abnormal post-accident vehicle pose changes cause a rise in accident confidence, demonstrating the models’ sensitivity to traffic anomalies.  

Finally, in Figure \ref{Visualization} (d), after the traffic light turns green, the white car begins moving forward while the car ahead remains stationary. The distance decreases, relative speed increases, and semantically there appears to be a rear-end risk. In reality, however, the distance remains safe but is obscured by partial occlusion. Our model mistakenly raises an accident warning, while the baseline remains at a moderate-to-low confidence level in the multi-vehicle scenario and is not misled.

\section{Conclusion}
This research presents an accident anticipation pipeline that leverages a dynamic graph convolutional network enriched with geometric and semantic cues. To support systematic evaluation, we also introduce the Multi-source Accident Anticipation dataset, which contains diverse accident scenarios with comprehensive annotations. In addition, we design a world model-based video augmentation pipeline that generates synthetic driving scenes and verify its benefit in enhancing anticipation accuracy. Extensive experiments have proven that our method is superior to previous methods on several benchmarks, setting new performance standards. Importantly, our analysis highlights that synthetic data, although still imperfect, provides measurable benefits for targeted tasks. Looking ahead, future work should emphasize improving the realism of generated content, particularly in terms of object consistency, noise suppression, and scene fidelity.

\section*{Acknowledgment}
This work was supported by the Science and Technology Development Fund of Macau [0122/2024/RIB2, 0215/2024/AGJ, 001/2024/SKL], the Research Services and Knowledge Transfer Office, University of Macau [SRG2023-00037-IOTSC, MYRG-GRG2024-00284-IOTSC], the Shenzhen-Hong Kong-Macau Science and Technology Program Category C [SGDX20230821095159012], the Science and Technology Planning Project of Guangdong [2025A0505010016], National Natural Science Foundation of China [52572354], the State Key Lab of Intelligent Transportation System [2024-B001], and the Jiangsu Provincial Science and Technology Program [BZ2024055].

\section*{DATA availability}

The data used in this work are publicly released at:  
https://github.com/humanlabmembers/Multi-source-Accident-Anticipation.

\backmatter

\bibliography{sn-bibliography}

\end{document}